\documentclass{article}

\usepackage{microtype}
\usepackage{graphicx}
\usepackage{subfigure}
\usepackage{booktabs} 

\usepackage{hyperref}
\usepackage{graphicx}
\usepackage{bbm}
\usepackage[utf8]{inputenc} 
\usepackage[T1]{fontenc}    
\usepackage{hyperref}       
\usepackage{url}            
\usepackage{booktabs}       
\usepackage{amsfonts}       
\usepackage{nicefrac}       
\usepackage{microtype}      
\usepackage{amsmath,tabu}
\usepackage{blkarray}
\newcolumntype{L}{>{$}l<{$}}
\DeclareMathOperator*{\argmax}{argmax}

\usepackage{float}
\newcommand\mydots{\hbox to 1em{.\hss.\hss.\hss}}

\usepackage[accepted]{icml2019}
\icmltitlerunning{Constraint Satisfaction Propagation: Non-stationary Policy Synthesis for Temporal Logic Planning}

\begin{document}

\twocolumn[
\icmltitle{Constraint Satisfaction Propagation:\\ Non-stationary Policy Synthesis for Temporal Logic Planning}



\icmlsetsymbol{equal}{*}

\begin{icmlauthorlist}
\icmlauthor{Thomas J. Ringstrom}{to}
\icmlauthor{Paul R. Schrater}{to,goo}
\end{icmlauthorlist}

\icmlaffiliation{to}{Department of Computer Science, University of Minnesota, Minneapolis, MN}
\icmlaffiliation{goo}{Department of Psychology, University of Minnesota, Minneapolis, MN}

\icmlcorrespondingauthor{Thomas J. Ringstrom}{rings034@umn.edu}

\icmlkeywords{Machine Learning, ICML}

\vskip 0.3in
]



\printAffiliationsAndNotice{}  

\begin{abstract}
Problems arise when using reward functions to capture dependencies between sequential time-constrained goal states because the state-space must be prohibitively expanded to accommodate a history of successfully achieved sub-goals.  Also, policies and value functions derived with stationarity assumptions are not readily decomposable, leading to a tension between reward maximization and task generalization.  
We demonstrate a logic-compatible approach using model-based knowledge of environment dynamics and deadline information to directly infer non-stationary policies composed of reusable stationary policies. The policies are constructed to maximize the probability of satisfying time-sensitive goals while respecting time-varying obstacles.  Our approach explicitly maintains two different spaces, a high-level logical task specification where the task-variables are grounded onto the low-level state-space of a Markov decision process.  Computing satisfiability at the task-level is made possible by a Bellman-like equation which operates on a tensor that links the temporal relationship between the two spaces; the equation solves for a value function that can be explicitly interpreted as the probability of sub-goal satisfaction under the synthesized non-stationary policy, an approach we term {\em Constraint Satisfaction Propagation (CSP)}.
\end{abstract}

\section{Introduction}
Natural tasks often involve sequential dependencies between individual goals which comprise a larger composite goal.  Imagine, for instance, a detective trying to solve a crime where they need to collect evidence from witnesses and arrest suspects for interrogation.  The detective will need to reason about the order in which these sub-goals are executed and may need to use knowledge of individual deadlines to put constraints on the possible sub-goal sequences.  For example, the detective knows that two key witnesses will be leaving town for work in the morning and the two main suspects will likely leave town later in the day. The detective will thus conclude that the witnesses must be questioned first so that there is enough time and evidence to arrest and interrogate the suspects, as they cannot be held in custody for longer than a day. The order in which the two witnesses are questioned and the order in which the two suspects are arrested does not matter for the satisfaction of the task which only requires that all sub-goals are met before their individual deadlines, leading to four distinct possible sequences of sub-goals that can be executed. Furthermore, the difficulty of this task is compounded by the fact that the detective must have knowledge of the underlying movement constraints and knowledge of the dynamics of the environment. If the detective knows that traffic will block particular routes during periods of the day, then a plan will need to anticipate these dynamic obstacles.
\begin{figure*}[!ht]
\centering
  \includegraphics[width=\linewidth]{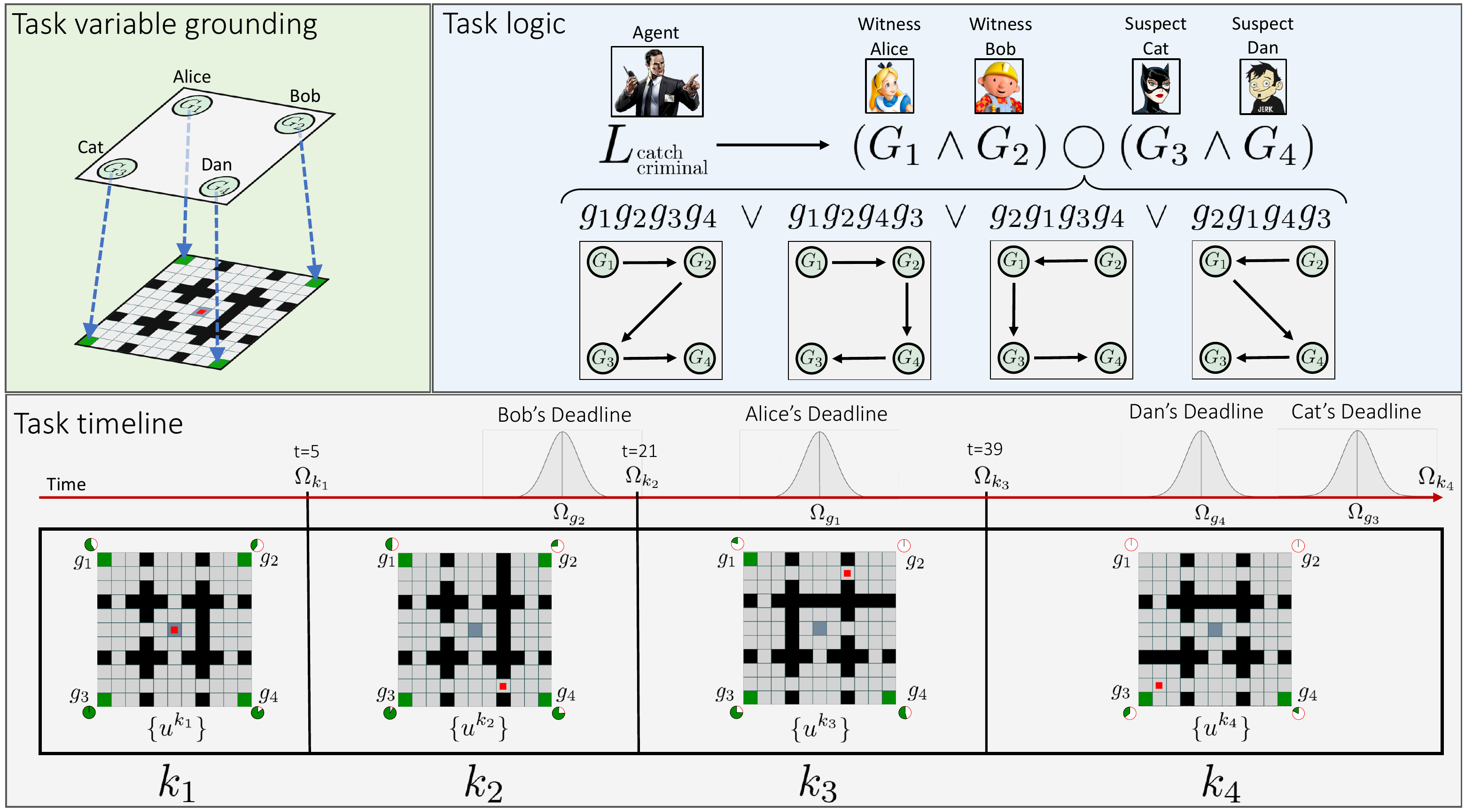}
\caption{\textbf{Problem Example}: You are a detective tasked with catching a bank robber.  You have the addresses of two witnesses to the crime Alice and Bob, but they are only home for a certain length of time before they leave town for work. The whereabouts of the suspects, Cat and Dan, are also known and they can be tracked down later but they are suspected to flee the city sometime in the evening. The witnesses must be questioned first (in any order) before arresting the suspects (in any order) to acquire evidence for interrogation.  You must plan for time-varying blockages in your route in order to complete the composite task on time.  \textbf{Green Panel}: Task variables assigned to people are grounded into the low-level state-space. \textbf{Blue panel}: The task logic $L$ is given and implies four different task graphs with their associated goal strings which would satisfy $L$ if all elements of the string returned true.  $G_1 \land G_2$ implies sub-strings of $g_1g_2$ or $g_2g_1$.  The "next" operator $\bigcirc$ concatenates strings together.  See \textit{Figure} \textit{\ref{fig:taskexample}} for the full list of identities for producing a disjunctions of strings.  \textbf{Grey Panel}: A depiction of the task with goal deadlines and transitions between world environments.  Each of the $K$ environments has an associated set of optimal controlled dynamics ${u^k}$ that transport the agent from state $x_i$ to target state $x_j$.  Each environment transitions to its successor after the time indicated on the timeline.  
}
  \label{fig:TaskGrounding}
\end{figure*}

Problems like the one faced by the detective are difficult to solve in standard reinforcement learning.  In the tabular domain, the difficulty arises from the necessity of expanding the state-space to be able to keep track of a trace (history) of successfully achieved sub-goal states.  In Deep RL the difficulties arise because there are effectively an almost limitless number of states due to the large dimensionality of the input space, thereby making it challenging to impose a notion of sequential goal dependency on the structure of the neural network architecture in addition to the difficulties of extracting sub-tasks from the network after end-to-end training.  In deep RL, recent methods such as RUDDER and TVT approach the problem of long term planning and reward-to-action credit assignment by using a recurrent network to use predictions about future rewards in order to rewrite the reward function \cite{hung2018optimizing, arjona2018rudder}.  Hierarchical approaches in the tabular domain have been successful in controlling over large sets of individual tasks using a compositional linear strategy, and have used these strategies for decomposing complex problems by inferring sub-goal representations embedded in a complex reward function  \cite{jonsson2016hierarchical,saxe2017hierarchy,earle2017hierarchical}. Still, current approaches in both domains don't constrain the sub-problems to be compatible with flexible planning under non-stationary rewards.  

Hierarchical RL (HRL) holds the promise of overcoming these complexities, but current theory is still insufficiently powerful to tackle the detective's problem.  HRL employs a few main strategies: breaking large markov decision processes into small sub-MDPs (MAXQ) \cite{dietterich2000hierarchical}, coarse-graining actions into temporally extended actions (Options) \cite{sutton1999between, precup2000temporal}, and coarse-graining states (state aggregation) \cite{bertsekas2005dynamic, singh1995reinforcement}. However, these policy decomposition strategies are seldom used to improve goal representation and are often motivated by reducing computational costs of policy learning. Moreover, these methods typically do not address a key characteristic of complex goals: time-constrained sequential dependencies.

While hierarchical approaches most often focus on policy structure, the basic assumptions made about reward functions fundamentally limit our ability to represent complex tasks.  Reward functions used in RL are usually not time-dependent, and many problems that should naturally be framed as non-stationary finite horizon problems are cast as infinite horizon discounted problems where the discount factor either guarantees convergence or represents agent preference.  Reward functions defined on the low-level state-space do not represent the sequential \textit{structure} of a task, and in the case of Deep RL, if the reward function is not a good proxy for the task it can result in "reward hacking," where the agent subverts the engineer's intent by finding clever but undesirable solutions to the task \cite{amodei2016concrete}.  However, in order to encode a task with long term sequential dependencies, one would need to expand the state-space to represent a trace of previously visited goal states such that the reward payout is conditioned on these visitations.  This state-space expansion would exponentially increase the computational complexity of the solution, which leads us to ask: Can tasks be formulated in such a way that is conducive to optimization without state-space expansion?  We provide an affirmative answer to this question.

In general, a task can be thought of as a logical statement composed of \textit{constraints}, which are variables that condition the truth value of the task.  Constraint variables provide a flexible way to reason about the conditions for task satisfaction, and a representation of a task can be constructed in a compositional manner using logical operators. The algorithm we propose, \textbf{Constraint Satisfaction Propagation} (CSP), is analogous to a Bellman value recursion \cite{bellmandynamic} in that it uses a backwards recursion to propagate information for constructing a control law. It differs, however, in that the information represented by the value function can be explicitly interpreted in terms of the probability of satisfying an individual constraint.  A set of value functions derived for each individual constraint variable are then used with their associated component policies in a forward propagation to compute the probability of solving the task under a composite non-stationary task-policy.  

Because CSP computes task-interpretable value functions, it provides a principled way of avoiding reward-based value function decomposition.  With CSP, we develop an approach to handling temporal constraints and goal dependencies by constructing a multi-tiered architecture that combines top-down logical task deduction and bottom-up policy inference to synthesize a non-stationary policy composed of members from a stationary policy ensemble.  The policies are synthesised to maximize the probability of satisfying a task specification. This approach bridges two spaces, one of which is a low-level state-space paired with a set of precomputed optimal stationary policies each controlling to a unique individual target state, and the other is an abstract task variable space that encodes sequential dependencies of individual sub-goal states and their deadlines.

To bridge these spaces, we compute an object called the \textbf{reachability tensor}.  This tensor plays a fundamental role in the communication between the two spaces by relating a policy's forward time dynamics to deadline distributions.  By committing up front to computing an ensemble of shortest-path policies, we can use their associated time-to-go representations to compute the reachability tensor. With the tensor we can effectively compute a composite non-stationary policy that satisfies a task specification via policy inference using a \textbf{Reachability Bellman Equation}.  The value function associated with the reachability Bellman equation, $\kappa_g^k(x_i|\Omega_c)$, hereafter called a \textbf{feasibility function}, can be explicitly interpreted as the probability of reaching a goal state $x_g$ from $x_i$ before deadline $\Omega_c$ while controlling optimally starting from the beginning of environment $k$. We consider dynamic worlds that have time-varying obstacles which remain constant for a given period of time which defines an environment. 

The compositionality of this algorithm is two-fold: firstly, by precomputing a policy ensemble across members of the set of environmental constraints, we can reuse them  for any temporal ordering of the environments to compute a set of feasibility functions $\{\kappa_g\}_{g\in \mathcal{G}}$ for each goal. Secondly, with this feasibility function set we can compute the probability of satisfying \textit{any} task specification under the environment dynamics and goal deadlines. The structure of this architecture induces a flexible bidirectional mapping between the space of temporal goals and the space of dynamics, which is made possible by virtue of the fact that the feasibility function can be explicitly interpreted in terms of the satisfaction of task variables.  

The major contribution of this paper is demonstrating that the reachability tensor is a crucial object that enables this compositional property because it is derived using the temporal dynamics information from both spaces.  With CSP, hierarchical problems can be solved with a combinatorics approach instead of reward maximization due to the fact that the feasibility function locally stores explicit information about the global connectivity of time-sensitive task variables defined on a dynamic graph under a non-stationary policy.  


\subsection{Related Work}

\subsubsection{Task Specification and Decomposition with Temporal Logic}
Task transfer has emerged as a significant problem in reinforcement learning, particularly when tasks have divergent objectives. Differences in objective functions can produce difficult to predict changes in policies, obscuring how tasks are related and motivates theory of policy and value function transfer between MDPs \cite{wilson2007multi}. A fundamental solution to this problem is to represent tasks within a common task space separate from the low-level state-space by decomposing tasks using a set of operators. 
Recently, temporal logic has proven increasingly successful at providing such a principled task space, which allows stationary, non-stationary, and complex objectives to be represented in a common space \cite{littman2017environment, li2017reinforcement, aksaray2016q}.  In addition, rich task descriptions can be inferred from demonstrations using temporal logical formulas \cite{shah2018bayesian}.  Such representations have been adapted to RL, and a connection between temporal logic and finite state machines has been exploited to provide a natural dynamic reward representation of a task \cite{icarte2018using}. Task transfer benefits from compositional goal representations through identifying common components. If the task decomposition has a congruent policy decomposition, the task compositionality of one problem can help structure the search for a solution to a new task with minimal recomputation.  Compositionality as a principle is of theoretical interest because of its combinatorial and efficient use of simple primitives \cite{lake2015human}. In the case of CSP the learning system has task transfer to the extent that components of the algorithm (ranging from low-level policy tensors to high-level feasibility functions) are exchangeable across different \textit{classes} of task specifications.

\subsubsection{Policy Representation and Computation}
While many hierarchical approaches can be interpreted as temporally extended action, Precup and colleagues \citeyearpar{sutton1999between, precup2000temporal} developed a general framework for controlling with temporal abstractions using policies, termination conditions, and initiation sets together to define an "option".  Recent work on this front has aimed to efficiently learn a good set of policies and termination conditions \cite{bacon2017option, achiam2018variational, liu2017eigenoption}. We use an options-like framework based on a set of stationary policies, where the termination condition is dictated deterministically by changes in the environment and the initiation set is unrestricted. While the options framework can be quite general, our approach restricts the allowed policies to a family of shortest-path stationary policies that have one absorbing terminal state.  This restriction is crucial for extracting a policy's joint time and state prediction necessary for constructing a reachability tensor.  One of the formulas used to compute this (eq. \ref{meanttg}) can be considered a "successor representation" (SR) \cite{dayan1993improving} with restrictions on the form of the policy and reward function. The SR has recently been used to learn good options \cite{machado2017eigenoption}, and \citealt{stachenfeld2017hippocampus} provide a compelling argument that the SR is encoded by place field firings in the hippocampus (for review see \citealt{gershman2018successor}). CSP is compatible with this perspective, and we suggest that the additional representations put forth could lead to new insights for understanding task generalization in computational neuroscience.

Reachability properties of control systems have also been studied in various contexts.  \citealt{abate2008probabilistic} first formulated the dynamic programming reachability recursion for analyzing the safety properties of hybrid systems. Also, \citealt{horowitz2014compositional} demonstrated a compositional method for Linear Temporal Logic (LTL) control using constrained reachability problems where goals are time-homogeneous and without deadlines. Recently, \citealt{haesaert2018temporal} showed how to solve time-homogeneous (stationary) policies satisfying LTL specifications in continuous space using reachability recursions.  To our knowledge, our work provides the first non-stationary policy synthesize for dynamic environments with time-constrained goals by leveraging the compositional property of feasibility functions.

\section{Base-level Policy Representation: LMDPs }
Since our algorithm relies on an ensemble of precomputed representations, we review the main theoretical tools used to compute it.  Our algorithm of choice for solving for a controlled transition matrix is the linearly solvable Markov decision process (LMDP) \cite{todorov2009efficient} due to its computational efficiency, however, the algorithm we develop is agnostic to the choice of Markov decision process (MDP) optimization method for obtaining the controlled dynamics.

An LMDP is an entropy-regularized relaxation of a standard MDP, defined as a three-tuple $(X,P,q)$ where $X$ is the set of discrete states, $P$ is an uncontrolled "passive" dynamics transition matrix $P : X \times X \rightarrow [0, 1]$, and $q$ is a cost function $q: X\rightarrow \mathbb{R}$. Whereas the standard Bellman equation for an MDP \cite{puterman2014markov} is defined as:
\begin{gather}
v^*(x) = \min_{a}\Big[g(x,a) + \mathop{\mathbb{E}}_{x'\sim T(x'|x,a)}[v^*(x')]\Big]
\label{eq:valueFunction}
\end{gather}
the LMDP redefines the loss function $g(x,a)$ as $q(x) + KL[u(x'|x)||p(x'|x)]$, where the KL-divergence is between the  passive dynamics and controlled dynamics (policy).  The optimization is now directly over the dynamics itself, rather than action variables. For LMDPs, $u(x'|x)$ is considered to be a policy, not to be confused with an MDP policy $\pi(a|x)$, and we adopt this nomenclature, but also refer to $u$ as the "controlled dynamics". Any MDP can be embedded into an LMDP by converting $g(x,a)$ and $T(x'|x,a)$ into $q(x)$ and $p(x'|x)$ by solving a small system of linear equations for each state, however, typically if one uses a constant action cost the passive dynamics from one state will be a uniform distribution over reachable next-states.  

Substituting in the new loss function gives rise to a discrete linear Bellman equation
\begin{gather}
v^*(x) = \min_{u(x'|x)} \Big[q(x) + KL(u||p) + \mathop{\mathbb{E}}_{x'\sim u}[v^*(x')]\Big]
\label{eq:original}
\end{gather}
The action variable has been replaced with a controlled transition matrix $u(x'|x)$ and we would like to find the optimal policy
$u^*(x'|x)$ that minimizes the new loss function.  By exploiting a transform on the value function which converts it to a \textit{desirability} function $z(x) = exp(-v(x))$, it can be shown that the optimal controlled dynamics can be computed by rescaling the passive dynamics by the optimal desirability function, which is the largest eigenvector of $QP$
\begin{gather}
u^*(x'|x) = \frac{p(x'|x)z(x')}{G[z](x)}\\
\mathbf z = QP\mathbf{z}
\label{eq:original}
\end{gather} 
where $Q$ is a diagonal matrix with elements $Q_{ii} = exp(-q(x_i))$ and $P$ is the passive transition matrix.  The rescaling is normalized by $G[z](x) = \mathop{\mathbb{E}}_{x\sim p}z(x)$.

\section{Constraint Satisfaction Propagation}
We introduce a new algorithm called Constraint Satisfaction Propagation which computes a composite non-stationary task policy from a precomputed ensemble of stationary policies.  The composite policy is synthesised to traverse one of many possible valid sequences of deadline-constrained goals states $x_g$ that satisfies a logical statement $L$.  These problems are defined on worlds with time-varying obstacles that remain constant for an period of time, which we call an \textit{environment}, indexed by $k$.  Synthesizing this non-stationary policy requires us to relate the time-dynamics of the low-level controller to the deadlines of the high-level task variables. This is made possible by computing a reachability tensor, $R$, which encodes the probability an agent can reach a state before a deadline.  With this tensor we can compute feasibility functions, $\kappa_g^k(x_i)$, which encodes the probability of satisfying a grounded sub-goal $h(G_c)\rightarrow x_g$ before its deadline under a component non-stationary policy.  Then, with the set of feasibility functions and their associated non-stationary policies, we can compute a composite task-policy using a forward-time propagation.


We begin this section by describing how to define LMDPs to create an ensemble of controlled dynamics to each state in an environment (\ref{polensemble}), followed by sections (\ref{deadlines}) and (\ref{logicsection}) which contain a descriptions of the deadlines and task logic we use for specifying tasks. Then we show how to create the reachability tensor (\ref{reachtensor}) with the deadlines for making temporally extended predictions. In (\ref{backwardrecursion}), we discuss how this tensor is acted on by a backwards recursion, the Reachability Bellman equation, to produce feasibility functions which summarize the temporal relationship between the low-level dynamics and the satisfaction of task variables.  Lastly, in (\ref{Forward}), we will introduce a forward pass algorithm for the task logic evaluation and synthesis of the non-stationary task policy $\pi_{L,x_0}$.

\subsection{Policy Ensemble} \label{polensemble}
We compute an ensemble of polices where each policy is the solution to an LMDP with a $Q$-matrix stored in the tensor $Q_{x,x'}^{k,j}$. Each $Q$-matrix is constructed by uniquely pairing a set of obstacle states with an individual target state in $X$. That is, if there is a set $\mathcal{B}=\{B_k\}_{1:K}$ containing obstacle states $B_k \subset X$ for each environment, then for each member $(B_k, x_j)$ of the Cartesian product $\mathcal{B} \times X$ there is a $Q$-matrix and a corresponding optimal policy $u_{\rightarrow j,k}^*(x'|x)$ where $j$ indexes the single target state in $X$.  Henceforth we will drop the $*$ notation and $u_{\rightarrow j}^k(x'|x)$ will be used to denote the optimal controlled dynamics.  Each $Q$-matrix giving rise to a policy is defined as:
\begin{gather}
Q_{i,i}^{k,j}=\begin{cases}
    0, & \text{if $x_i \in B_k$}\\
    1, & \text{elif $x_i \equiv x_j$}\\
    \epsilon, & \text{otherwise}\\
 \end{cases}
 \quad
 \xrightarrow[P]{LMDP}\quad u_{\rightarrow j}^{k}(x'|x)
\end{gather}
Since the elements of $Q$ are negatively exponentiated LMDP cost functions $q(x)$, the obstacles are given infinite costs and the target state has a cost of zero.  All other internal states $x_I$ have a constant cost $q(x_I) = -log(\epsilon)$ for $\epsilon \in (0,1)$.
Each $u_{\rightarrow j}^k(x'|x)$ is a shortest path LMDP solution for reaching $x_j$ while avoiding obstacle states $B_k$, and the LMDP is solved using a passive dynamics matrix $P$ derived from an MDP transition tensor with four actions corresponding to movement in the cardinal directions and one null action. The index $j$ will always denote the absorbing target state and as a reminder we use $\rightarrow j$ as long-form notation, however, we will sometimes drop the arrow when necessary for compactness.  We organize all $Q$ matrices into a tensor $Q_{x,x'}^{k,j}$ and each matrix $Q_{:,:}^{k,j}$ corresponds to a slice of the controlled transition tensor $U_{:,:}^{k,j}$.

\subsection{Deadline Distributions} \label{deadlines}
We assume that the agent has knowledge of the environment and goal deadlines ($\Omega_k,\Omega_c)$ in the form of a probabilistic model.  Environment transition deadlines are required to compute where the agent can go for the duration of an environment's period, and goal deadlines are required to compute if the objective can be achieved during a period.
The PDFs $f_{\Omega_{g}}(t)$ and $f_{\Omega_{k}}(t)$ are the deadline distributions associated with a goals $G_c^{\Omega_c}$ and environment change points.  We define periods by environment start and end times $d_k$ and $d_{k+1}-1$. For this paper the environment transition model is deterministic and the goal deadline model is probabilistic.
\begin{figure*}[t]
\centering
  \includegraphics[width=\linewidth]{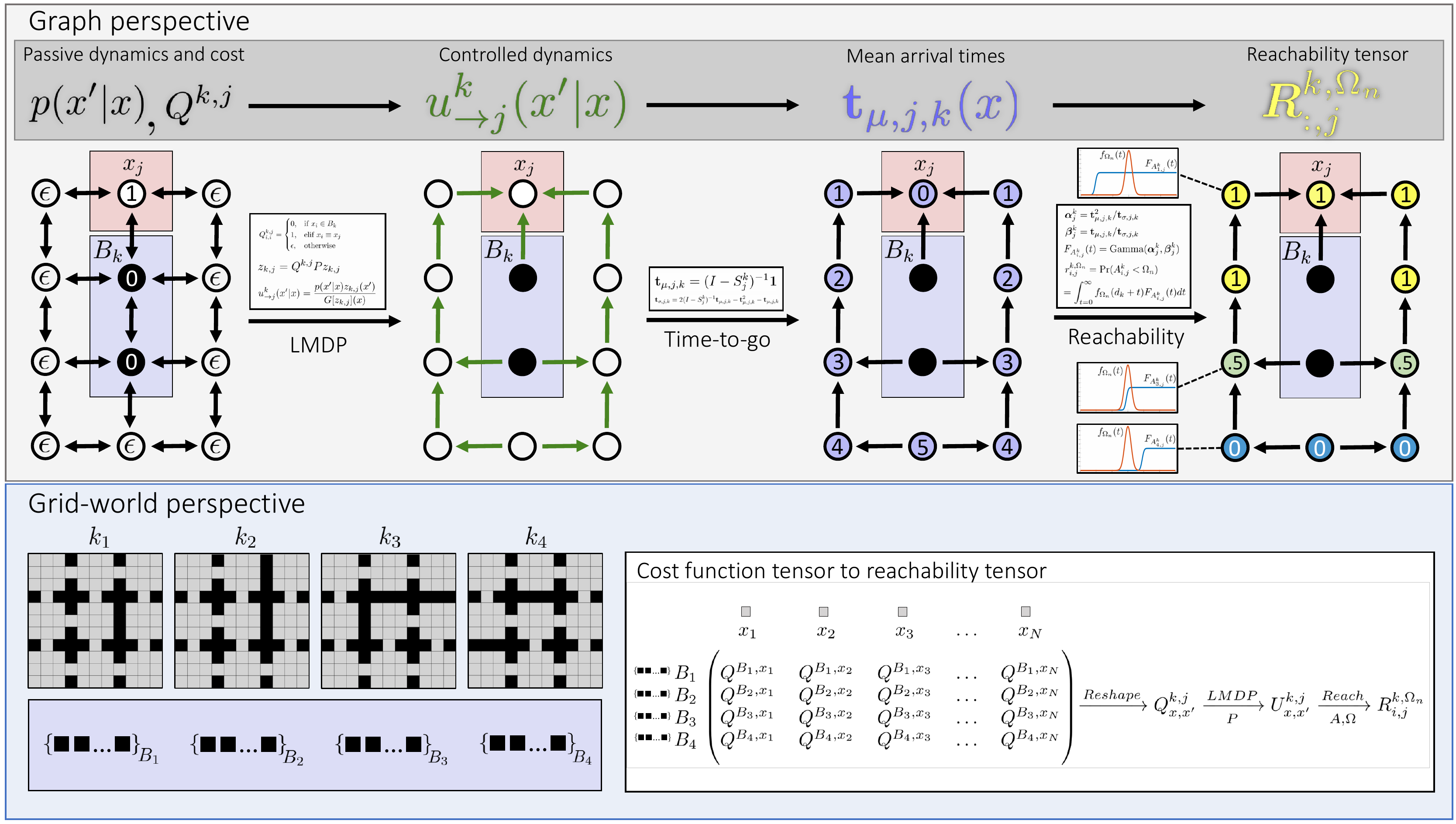}
\caption{The stages of producing a reachability tensor $R$.  (Top) The perspective from the point of view of the underlying graph showing how one LMDP cost matrix, $Q^{k,j}$, defined by target state $x_j$ and obstacle state set $B_k$, induces a member of $R$ with a deadline distribution $\Omega_n \in \{\Omega_k, \Omega_c\}$.  (Bottom) Four obstacle state sets on a gridworld are paired with each $x_j$ to create a cost tensor $Q$, low-level policy tensor $U$, and $R$.  This process is summarized by the bottom left panel of Figure \ref{fig:taskexample}.}
\label{fig:buildreach}
\end{figure*}
\subsection{A Simple Temporal Logic} \label{logicsection}
The temporal logic we use in this paper is tailored to our problem domain and is simpler than well known temporal logics like Linear Temporal Logic (LTL) \cite{pnueli1977temporal}.  LTL uses modal operators such as \textit{Until}, \textit{Always}, and \textit{Eventually}. We avoid modal logic in order to focus on policy synthesis for satisfying goal sequences.  Integration of CSP into LTL will be addressed in future work.  

\begin{figure*}[!h]
  \includegraphics[width=\linewidth]{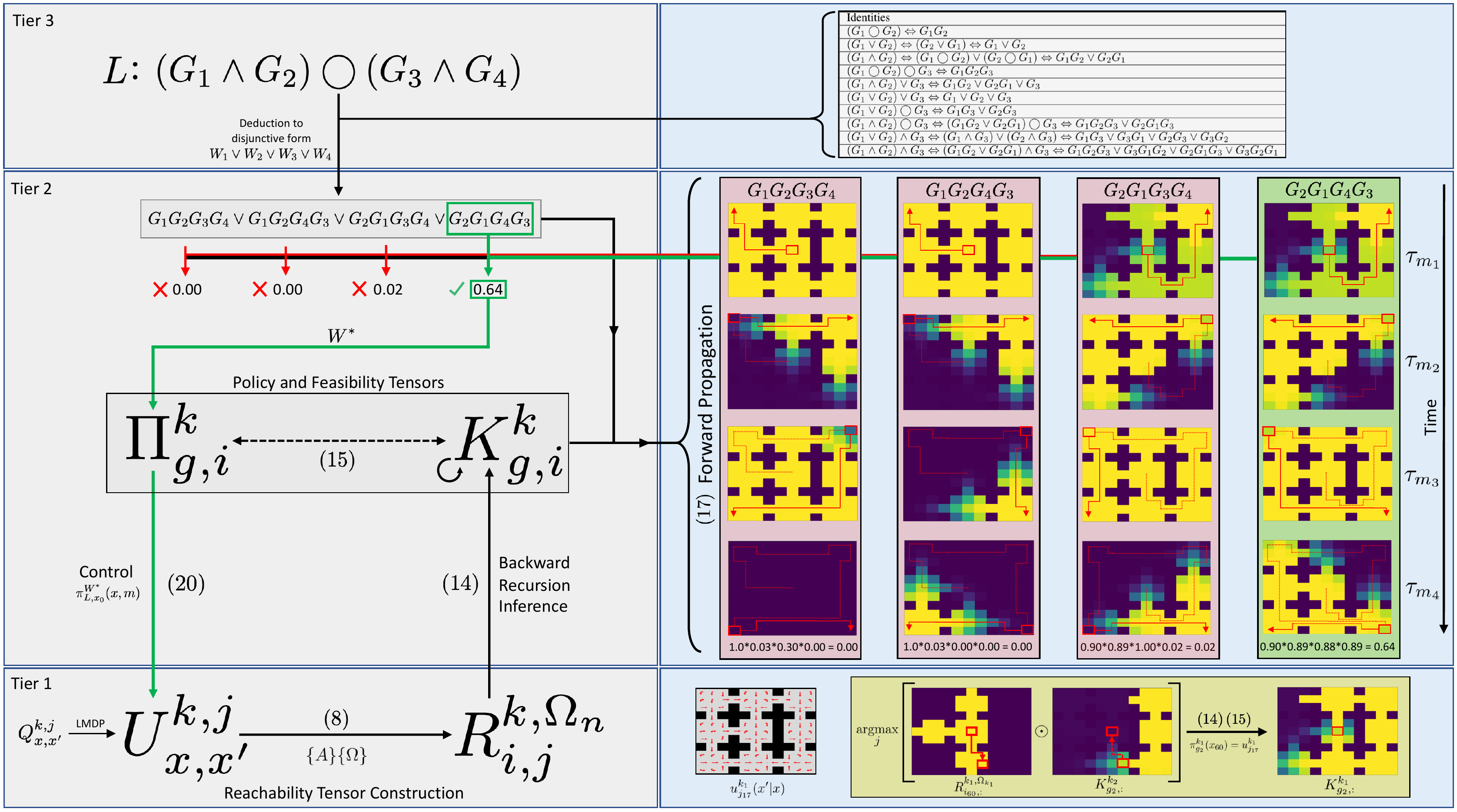}
  \caption{Tier 1 (Grey) contains the precomputed tensors from Figure \ref{fig:buildreach}. A cost tensor $Q$ induces policy tensor $U$ through LMDP optimization.  The reachability tensor $R$ is created with time-to-go information from $U$ and deadline variables.  Tier 1 (Blue) shows a single policy from a slice of $U$, and an example (yellow panel) of the feasibility computation for a single state $x_{60}$ (red box) that demonstrates how the reachability tensor is used to calculate the feasibility function for $x_{60}$, resulting in a policy selection for period $k_1$.  Tier 2 (Gray) contains feasibility and policy tensors.  Tier 2 (Blue) shows how the feasibility tensor and disjunctive logic from Tier 3 are combined to compute the probability of task $L$ defined by the a set disjunction of composite goal sequences.  Each slice $K_{:,g}^k$ is a feasibility map over the state-space. Red boxes at state $x_i$ denote the value of $\kappa_{i,g}^k$.  The color intensity inside the box indicates the maximum probability (yellow = $1.0$, purple = $0.0$) of successfully reaching goal state $x_g$ by following the policy $\pi_{k,i}^g$ (solid red line) and contributes to the total probability of the word being true. Dashed red lines represent trajectory history from previous periods.  The highest probability goal sequence is chosen and indexes into the policy tensor $\Pi$ which directly controls low-level dynamics by setting the $j$ and $k$ index of $U$ (Green Arrows). }
  \label{fig:taskexample}
\end{figure*}

We use operators ($\land,\lor,\bigcirc$) over goal variables, $G_c^{\Omega_c}$ in order to define a task $L$. Goal variables are superscripted by their deadlines, but we will often drop this notation for simplicity. $\lor$ and $\land$ are OR and AND operators.  The statement $(G_1$ $\land$ $G_2)$ means $G_1$ and $G_2$ both must get done in any order before their deadlines. The $\bigcirc$ operator denotes "next," and enforces sequential constraints between goals by concatenating variable strings together. This means that the goal arrival times for variables in $G_{\text{1}}\bigcirc G_{\text{2}}$ must satisfy, $\tau_{g_\text{1}}<\tau_{g_\text{2}}$. A logical statement $L$ always reduces down to a disjunctive form of ORs over goal variable sequences (words), $L \implies W_1 \lor \dots \lor W_n$ where $W_i = G^{1}G^{2}\mydots G^{M_i}$.  Figure \ref{fig:taskexample} shows a list of identities for reducing a statement down to a disjunctive form.  

All goal variables are grounded onto a member of $X$, $\forall G_{c} \in \mathcal{G}$ : $h(G_{c})\rightarrow x_{g_c}$, and a controller will return a state-time certificate $a_{x_i,\tau} = (x_i,\tau)$ only when it passes though a goal state.  Note that $g$ is an index for a member of $X$, and $c \in \{1,2, \mydots C\}$ indexes both the goal variable and the index corresponding to its grounding\footnote{Example: For $h(G_2)= x_9$, $c=2$, $g=9$, $h^{-1}(x_{9})=G_2$ we can reference the grounding state with $x_{g_2} \equiv x_9$}.  A controller-produced word $w_{\pi} = a^{1}a^{2}\mydots a^{M}$ is evaluated against $W$, which returns true if the certificate goal order matches and the deadlines are satisfied:
\begin{gather}
    p(W_i=True|w_{\pi}) = \prod_{m=1}^M\mathbbm{1}_{W_i(m)}(w_{\pi}(m))\label{wordeval}\\
    \mathbbm{1}_{G_c^{\Omega_c}}(a_{x_i,\tau}) = 
    \begin{cases}
    1 & \text{if} \text{\,\,}(h^{-1}(x_i) \equiv G_c) \land (\tau \leq \Omega_c)\label{wordindicator}\\
    0 & \text{otherwise}
 \end{cases}
\end{gather}
Here, $W_n(m)$ and $w_{\pi}(m)$ dereference the $m^{\text{th}}$ entry in strings $W$ and $w_{\pi}$.  

\subsection{Reachability Tensor} \label{reachtensor}
Planning with deadlines requires forecasting which low-level policy transitions will meet goal deadlines while avoiding time-varying path blockages from obstacles, a concept termed {\bf reachability}. We can compute the set of forecasts from the distribution of policy-conditional hitting times $A_{i,j}^k \sim f_{A_{i,j}^k}(t)$, where $A_{i,j}^k$ is a random variable of the time taken to transition from state $x_i$ to $x_j$ under $u_{\rightarrow j}^k(x'|x_i)$ (in number of steps for discrete time). Because our LMDP solutions have one target state we can model them as absorbing Markov chains, and compute the first two moments of the arrival time distribution for the target state $x_j$ from any initial state $x_i$. 

We define a reachability tensor $R_{i,j}^{k,\Omega_n}$ (Figure \ref{fig:buildreach}) that spans the set of reachability forecasts needed for planning. Each element $r_{i,j}^{k,\Omega_n}$ of $R_{i,j}^{k,\Omega_n}$ is the probability of reaching state $x_j$ from $x_i$ under the dynamics of $u_{\rightarrow j}^k$ before deadline $\Omega_{n}$,
\begin{gather}
r_{i,j}^{k,\Omega_{n}} = \text{Pr}(A_{i,j}^k < \Omega_{n}) = \int_{t={0}}^{\infty} f_{\Omega_{n}}(d_k+t)F_{A_{i,j}^k}(t)dt
\label{reachfunction}
\end{gather}
where $d_k$ is the starting time of environment $k$, $f_{\Omega_{n}}$ is the deadline PDF, and  $F_{A_{i,j}^k}(t)$ is the cumulative distribution of policy-conditional hitting times. Deadline distribution $\Omega_n$ represents one of two distributions $\{ \Omega_k, \Omega_c \}$ which are the $k^{th}$ period deadline and a goal deadlines. We compute an approximation of the arrival-time CDF $F_{A_{i,j}^k}$ with the absorbing Markov chain formula for the first two moments of the arrival-time PDF of $u_{\rightarrow j}^k$ by using its submatrix $S_j^k$, with the $j^{th}$ row and column removed: 
\begin{gather}
\mathbf{t}_{\mu, j, k} = (I-S_j^k)^{-1}\mathbf{1} \label{meanttg}\\ \mathbf{t}_{\sigma, j, k} = 2(I-S_j^k)^{-1}\mathbf{t}_{\mu, j, k}-\mathbf{t}_{\mu, j, k}^2-\mathbf{t}_{\mu, j, k}
\end{gather} 
We model the cumulative distribution for arrival-times as a gamma CDF:
\begin{gather}
F_{A_{:,j}^k}(t) = \text{Gamma($\boldsymbol{\alpha}_j^k, \boldsymbol{\beta}_j^k$)}\\
\boldsymbol{\alpha}_j^k = \mathbf{t}_{\mu, j,k}^2/\mathbf{t}_{\sigma, j,k}\\
\boldsymbol{\beta}_j^k = \mathbf{t}_{\mu, j,k}/\mathbf{t}_{\sigma, j,k}
\end{gather} 

For this paper, we compute an $R$ which includes all discrete deterministic deadlines $\Omega_t$ through time $T$ plus all deadlines for individual goals variables $\Omega_c$, where an environment's deadline $\Omega_k$ is a member of the deterministic superset, $\{\Omega_k\}_{1:K} \subset \{\Omega_t\}_{1:T}$.  The reason we compute $R$ for all deterministic deadlines is because it is required by the forward propagation algorithm discussed in section \ref{Forward}.


\subsection{Backward Recursion: component feasibility functions from reachability recursion} \label{backwardrecursion}
The reachability tensor is the key object used in a recursive equation we term the \textbf{Reachability Bellman equation}.  This equation computes $\kappa_{g_c}^{k}(i)$ which represents the probability of reaching $x_{g_c}$ from state $x_i$ starting in period $k$ if the agent follows an associated non-stationary policy.  
\begin{eqnarray}
  \kappa_g^{k}(x_i) = \max_{j} \Big[ R_{i,j}^{k,\Omega_c}\mathbbm{1}_g(j) + R_{i,j}^{k,\Omega_{k}}\kappa_g^{k+1}(x_{j})\mathbbm{1}_{\bar{g}}(j)\Big]
  \label{eq:mainkappa}
\end{eqnarray}
The first term in the max function adds in the probability of being able to achieve the goal during period $k$, $\mathbbm{1}_g(j)$ and $\mathbbm{1}_{\bar{g}}(j)$ are indicator functions for $j = g$ and $j \neq g$ respectively.  The second term propagates the maximum future goal feasibility from non-goal states through the state-space for the preceding period. (Fig \ref{fig:taskexample}, Bottom Right).


The policy $\pi_g^{k}(x_i)$ directly specifies an intermediate "target" state $x^{k,g}_{j^*}$ that maximizes the future reachability of $x_g$ under the control of the LMDP solution indexed by $j^*$. For each goal $G_c$ and environment, we compute $\pi_{g_c}^{k}(x_i) = u_{\rightarrow j^*}^k$
\begin{gather}  
  x_{j^*}^{k,g}  =  \argmax_{j} \Big[R_{i,j}^{k,\Omega_c}\mathbbm{1}_g(j) + R_{i,j}^{k,\Omega_{k}}\kappa_g^{k+1}(x_{j})\mathbbm{1}_{\bar{g}}(j) \Big]\\
  \pi_g^{k}(x_i) = u_{\rightarrow j^*}^k(x'|x)
  \label{policy}
  \end{gather} 
The value functions are stored in a tensor $K_{g,i}^{k}$, and non-stationary policies are stored in the tensor $\Pi_{g,i}^{k}$ and point to an optimal $j^*$ which indexes into $U_{x_,x'}^{k_,j^*}$.  


\subsection{Forward Propagation: task-policy synthesis and evaluation with feasibility functions} \label{Forward}

Because component feasibility functions have a probabilistic interpretation for each goal variable, they can be used in a forward-time pass to compute the probability of satisfying a task, $L$, under a composite task-policy $\pi_{L,x_0}^{W}$.  $L$ defines a set of acceptable goal sequences $W_i = G^{1}G^{2}\mydots G^{M_i}$, where $M_i$ is the length of string $W_i$ and $N_W$ is the number of distinct words.  $L$ is assumed to be in disjunctive normal form across the set of these words, $L = \vee_{i=1}^{N_W} W_i$. Satisfying any word in a disjunction satisfies $L$, so we can evaluate the probability of task satisfaction by using the words as instructions for low-level policy indices.

The probability of task satisfaction is equivalent to evaluating the joint probability that all the sub-goals are reached and satisfy the temporal constraints. The primary reason for computing reachability is that it allows for the simple evaluation of the feasibility of sub-goal traces. We can compute word satisfiability by evaluating sub-goal traces with the $K$ tensor.  
Considering $L$ and $W$ as Bernoulli random variables, the estimate of the probability of the word being true under the set of CSP tensors $\mathcal{T}=\{\Pi, K, R\}$ is a product over feasibility functions, and the most feasible goal sequence is given by the most probable word $W^*$:
\begin{gather}
Pr(W_i=1|w_i\sim\pi_{L,x_0}^{W_i}) = \prod_{m=0}^{M-1} \hat{\kappa}_{g_{m+1}}^{k_{m}}(x_{g_{m}})\label{prwtrue}\\
\max_{\pi_{L,x_0}}[Pr(L=1)|\mathcal{T}]= \max_{i}[Pr(W_i=1)|\pi_{L,x_0}^{W_i}]\\
W^* = \argmax_{W_i}[Pr(W_i = 1)]
\end{gather}

Where $g_{m}$ is shorthand for the index $g$ from $h(W(m))=x_g$, and $x_{g_0}=x_0$ is the agent's starting state. Eq. \ref{prwtrue} is a stochastic version of eq. \ref{wordeval} which treats input characters produced by the policy as random random variables with an accepting condition given as a Bernoulli R.V. with parameter $\hat{\kappa}_{g_{m+1}}^{k_{m}}(x_{g_m})$.

The only complication in evaluating feasibility is when the sub-goal $W(m)$ changes to $W(m+1)$ in the middle of an environment's period at time $\tau_m$. The feasibility of $W(m+1)$ is not represented by $K$ for intermediate times, so now the feasibility depends on what can be accomplished in the remaining time, requiring a simple computation. We call these computations ``stitch" functions $\hat{\kappa}$ and $\hat{\pi}$, because they stitch in the additional feasibility information not accounted for by $K$. We compute $\hat{\kappa}_{g_{m+1}}^{k_{m}}(x_{g_{m}})$ over the remaining time of the period, $t_{m} = d_{k+1}-\tau_m$, with one step of eq.~\ref{eq:mainkappa}, using $r_{i=g_m,j}^{k_{m},\Omega_{t_m}}$ conditioned on the previous goal state. The arrival time $\tau_m$ can be stochastic, so we can condition on a time which is a high confidence bound estimated from the $99^{th}$ percentile of the hitting time distribution.  This distribution can be computed by convolving each of the discrete arrival distributions used between $g_m$ and $g_{m+1}$, or alternatively, as done for the examples in this paper, by the forward evolution of the system under the policy.  It is also possible to keep track of the state distribution over the entire word and marginalize the feasibility calculations over each sub-goal arrival time distribution.

The composite task policy calls (eq.\ref{policy}) over the most feasible goal sequence:  
\begin{gather}
\pi_{L,x_0}^{W^*}(x,m) \rightarrow \pi_{g_{m+1}}^{k}(x)\rightarrow u_{\rightarrow j^*}^k(x'|x)
\end{gather} 

At runtime, the agent follows and maintains its position $m$ in $W^*$ where transitions to $m'$ are governed by a transition matrix encoding $W^*$.

\section{Results}
\subsection{Theoretical Results: Complexity Comparison}

CSP tackles non-stationary problems with logical task-specifications that are conditioned on a trace (history) of sub-goal states. For this reason it becomes challenging to compare it against a benchmark algorithm for theoretical reasons.  Consider a comparable hypothetical problem to the one posed in this paper which we will call a trace-augmented LMDP (taLMDP) which would embed a CSP task $L$ into an LMDP. This LMDP needs to condition on the success/failure of its sub-goal history traces to compute its reward function payout, requiring an augmented state-space to represent the equivalent task of size $2^GN$, because there are $2^G$ distinct traces. Computing a non-stationary (stationary) policy for an LMDP has time-complexity of $\mathcal{O}(N^2T)$ ($\mathcal{O}(N^3)$) for a state-space of size $N$ and $T$ time steps. Therefore the taLMDP has a time-complexity of $\mathcal{O}(2^{2G}N^2T)$ which suffers from the curse of dimensionality.  

Additionally, the taLMDP has limited transfer learning capabilities and is unable to exploit the compositional structure of the task space. The analysis above is only with respect to a taLMDP embedding for a single task, $L$. However, the strength of CSP is that it requires no additional computation beyond the forward propagation if a new task $L'$ is formulated on the same environment sequence, goal variables, and deadlines.  On the other hand, a taLMDP value function will need to be recomputed for each $L'$.  Computing a set of taLMDPs for all possible $L$ adds another layer of intractability when scaling up these methods to complex structured reward functions.

To evaluate the computational complexity of CSP, the stages of the algorithm can be grouped into two types.  The first type is task-variable invariant which includes the computation of the policy basis and time-to-go representations, both of which are paid up-front once per domain (set of environments).  The second type is for the temporal logic control, which is a cost for each task specification that differs in deadline distributions. 

The complexity of computing a policy basis depends on the sophistication of the representation, but it amounts to computing a set of size $D$ of stationary policies, each with a cost $\mathcal{O}(N^3)$, for an overall complexity of $\mathcal{O}(DN^3)$. LMDP solutions can be computed with the power method or by solving a linear system \cite{todorov2009efficient}, so $\mathcal{O}(N^3)$ is an upper bound on the computation. If we compute a dense policy basis, we have $D=KN$ policies for a worst case complexity of $\mathcal{O}(KN^4)$.  The complexity of the arrival-time computations is similar, where the moments of hitting times are computed for each target state and environment, at cost of $N^3$, for an overall cost of $\mathcal{O}(KN^4)$. While this complexity is undesirable, in practice the policy computation benefits from sparcity, it also may be possible to significantly reduce the complexity of both policy and arrival-time computations using hierarchically structured LMDPs \cite{saxe2017hierarchy, jonsson2016hierarchical} and linear systems solvers tailored to sparse systems.  

The time complexity for the reachability tensor is $\mathcal{O}(GKN^2T)$, due to the fact that for every environment there are $N^2$ inner products of size $T$, and there are $G$ goal deadlines for $K$ environments. The feasibility function computation is a max operator over a point-wise multiplication which implies 
$\mathcal{O}(KGN^2)$. Lastly task evaluation is $\mathcal{O}(CGN)$ given that each string requires recomputing the feasibility function for the inter-period arrival times, which is one point-wise max operation of size $N$ for $G$ goals and $C$ strings.


\begin{figure}[!h]
  \includegraphics[width=\linewidth]{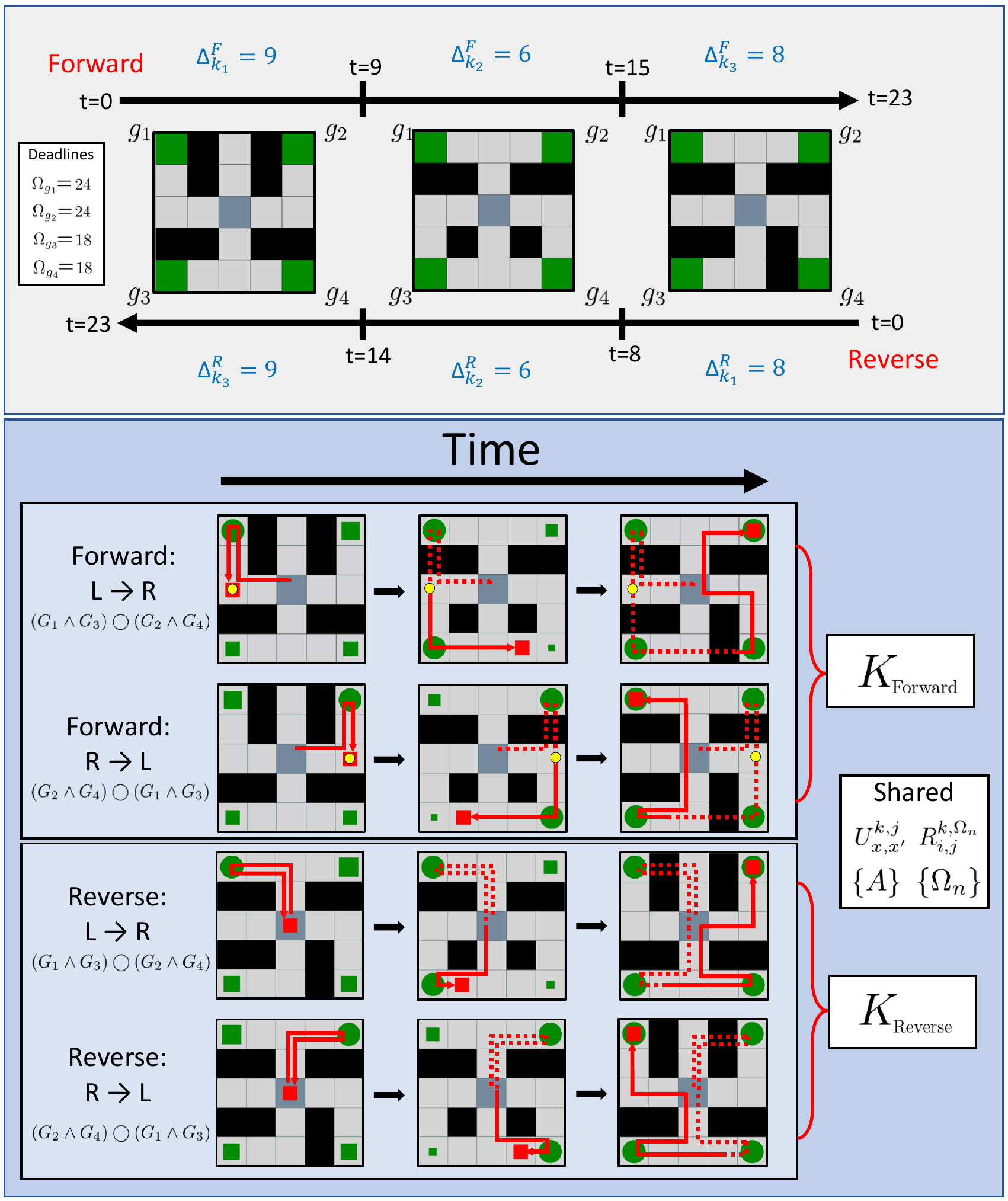}
  \caption{Compositionality Example: (Grey) Forward and Reverse environment sequences with shared goal deadlines and environment durations $\Delta_k$. (Blue) The agent (red square) is trailed by a solid red line for the path it traveled during the environment, and dashed lines represent its history. Green squares are the goals and the size indicates how much time is remaining before the deadline.  Green circles indicate that the goal was successfully visited.  Yellow circles indicate that the agent waited for a wall to open.}
  \label{fig:resultsfig}
\end{figure}
\subsection{Conceptual Demonstration}

In light of the theoretical comparison, we provide a proof of principle demonstration of the compositional efficiency of CSP. \textit{Figure} \ref{fig:resultsfig} (Grey) depicts two different environment progressions, one in which a sequence of three environments is played forward and one in which the order reversed.  Each individual environment has the same duration for both sequences, though this isn't a restrict requirement since they share an $R$ with all deterministic deadlines.  Because forward and reverse scenarios share environments, they have equivalent policy tensors, hitting times, goal deadlines, and reachability tensors.  The only thing that requires recomputation is the feasibility function which is computationally inexpensive.  (Blue) shows an example of two tasks for each environment sequence. In one the agent must visit the left two goal states first, and then the right two (in any order). In the second task the agent must visit the right two first, and then left. This example demonstrates that different tasks specified \textit{within} a forward or reverse sequence do not require recomputation of the feasibility tensor, the only computation needed is the evaluation of the different tasks.  However, this scenario isn't limited to only two tasks, all possible tasks entailed from the logical syntax share a feasibility tensor.  We provide code and video of the examples.\footnote{\url{https://github.com/constraintsatprop/CSP/wiki}}

\section{Conclusions}
We have demonstrated a new algorithm for synthesizing non-stationary policies to satisfy tasks with sequential goal constraints, deadlines, and time-varying obstacles.    By using temporal constraint logic to construct compositional task families, we enable the solution of complex, non-stationary problems in a hierarchical framework.  The approach provides a principled way to scale transfer learning to task families by ensuring congruence between objective and policy decompositions.  We believe that the architectural principles that we propose are necessary for tackling difficult planning problems, and that the principles can be enforced while scaling up the approach with more flexible state and policy representations.

\bibliography{references}

\begin{thebibliography}{32}
\providecommand{\natexlab}[1]{#1}
\providecommand{\url}[1]{\texttt{#1}}
\expandafter\ifx\csname urlstyle\endcsname\relax
  \providecommand{\doi}[1]{doi: #1}\else
  \providecommand{\doi}{doi: \begingroup \urlstyle{rm}\Url}\fi

\bibitem[Abate et~al.(2008)Abate, Prandini, Lygeros, and
  Sastry]{abate2008probabilistic}
Abate, A., Prandini, M., Lygeros, J., and Sastry, S.
\newblock Probabilistic reachability and safety for controlled discrete time
  stochastic hybrid systems.
\newblock \emph{Automatica}, 44\penalty0 (11):\penalty0 2724--2734, 2008.

\bibitem[Achiam et~al.(2018)Achiam, Edwards, Amodei, and
  Abbeel]{achiam2018variational}
Achiam, J., Edwards, H., Amodei, D., and Abbeel, P.
\newblock Variational option discovery algorithms.
\newblock \emph{arXiv preprint arXiv:1807.10299}, 2018.

\bibitem[Aksaray et~al.(2016)Aksaray, Jones, Kong, Schwager, and
  Belta]{aksaray2016q}
Aksaray, D., Jones, A., Kong, Z., Schwager, M., and Belta, C.
\newblock Q-learning for robust satisfaction of signal temporal logic
  specifications.
\newblock In \emph{Decision and Control (CDC), 2016 IEEE 55th Conference on},
  pp.\  6565--6570. IEEE, 2016.

\bibitem[Amodei et~al.(2016)Amodei, Olah, Steinhardt, Christiano, Schulman, and
  Man{\'e}]{amodei2016concrete}
Amodei, D., Olah, C., Steinhardt, J., Christiano, P., Schulman, J., and
  Man{\'e}, D.
\newblock Concrete problems in ai safety.
\newblock \emph{arXiv preprint arXiv:1606.06565}, 2016.

\bibitem[Arjona-Medina et~al.(2018)Arjona-Medina, Gillhofer, Widrich,
  Unterthiner, and Hochreiter]{arjona2018rudder}
Arjona-Medina, J.~A., Gillhofer, M., Widrich, M., Unterthiner, T., and
  Hochreiter, S.
\newblock Rudder: Return decomposition for delayed rewards.
\newblock \emph{arXiv preprint arXiv:1806.07857}, 2018.

\bibitem[Bacon et~al.(2017)Bacon, Harb, and Precup]{bacon2017option}
Bacon, P.-L., Harb, J., and Precup, D.
\newblock The option-critic architecture.
\newblock In \emph{AAAI}, pp.\  1726--1734, 2017.

\bibitem[Bellman(1957)]{bellmandynamic}
Bellman, R.
\newblock Dynamic programming.
\newblock \emph{Princeton University Press, Princeton, NJ}, 1957.

\bibitem[Bertsekas et~al.(2005)Bertsekas, Bertsekas, Bertsekas, and
  Bertsekas]{bertsekas2005dynamic}
Bertsekas, D.~P., Bertsekas, D.~P., Bertsekas, D.~P., and Bertsekas, D.~P.
\newblock \emph{Dynamic programming and optimal control}, volume~1, pp.\  319.
\newblock Athena scientific Belmont, MA, 2005.

\bibitem[Dayan(1993)]{dayan1993improving}
Dayan, P.
\newblock Improving generalization for temporal difference learning: The
  successor representation.
\newblock \emph{Neural Computation}, 5\penalty0 (4):\penalty0 613--624, 1993.

\bibitem[Dietterich(2000)]{dietterich2000hierarchical}
Dietterich, T.~G.
\newblock Hierarchical reinforcement learning with the maxq value function
  decomposition.
\newblock \emph{Journal of Artificial Intelligence Research}, 13:\penalty0
  227--303, 2000.

\bibitem[Earle et~al.(2017)Earle, Saxe, and Rosman]{earle2017hierarchical}
Earle, A.~C., Saxe, A.~M., and Rosman, B.
\newblock Hierarchical subtask discovery with non-negative matrix
  factorization.
\newblock \emph{arXiv preprint arXiv:1708.00463}, 2017.

\bibitem[Gershman(2018)]{gershman2018successor}
Gershman, S.~J.
\newblock The successor representation: its computational logic and neural
  substrates.
\newblock \emph{Journal of Neuroscience}, 38\penalty0 (33):\penalty0
  7193--7200, 2018.

\bibitem[Haesaert et~al.(2018)Haesaert, Soudjani, and
  Abate]{haesaert2018temporal}
Haesaert, S., Soudjani, S., and Abate, A.
\newblock Temporal logic control of general markov decision processes by
  approximate policy refinement.
\newblock \emph{IFAC-PapersOnLine}, 51\penalty0 (16):\penalty0 73--78, 2018.

\bibitem[Horowitz et~al.(2014)Horowitz, Wolff, and
  Murray]{horowitz2014compositional}
Horowitz, M.~B., Wolff, E.~M., and Murray, R.~M.
\newblock A compositional approach to stochastic optimal control with co-safe
  temporal logic specifications.
\newblock In \emph{IROS}, pp.\  1466--1473, 2014.

\bibitem[Hung et~al.(2018)Hung, Lillicrap, Abramson, Wu, Mirza, Carnevale,
  Ahuja, and Wayne]{hung2018optimizing}
Hung, C.-C., Lillicrap, T., Abramson, J., Wu, Y., Mirza, M., Carnevale, F.,
  Ahuja, A., and Wayne, G.
\newblock Optimizing agent behavior over long time scales by transporting
  value.
\newblock \emph{arXiv preprint arXiv:1810.06721}, 2018.

\bibitem[Icarte et~al.(2018)Icarte, Klassen, Valenzano, and
  McIlraith]{icarte2018using}
Icarte, R.~T., Klassen, T., Valenzano, R., and McIlraith, S.
\newblock Using reward machines for high-level task specification and
  decomposition in reinforcement learning.
\newblock In \emph{International Conference on Machine Learning}, pp.\
  2112--2121, 2018.

\bibitem[Jonsson \& G{\'o}mez(2016)Jonsson and
  G{\'o}mez]{jonsson2016hierarchical}
Jonsson, A. and G{\'o}mez, V.
\newblock Hierarchical linearly-solvable markov decision problems.
\newblock In \emph{ICAPS}, pp.\  193--201, 2016.

\bibitem[Lake et~al.(2015)Lake, Salakhutdinov, and Tenenbaum]{lake2015human}
Lake, B.~M., Salakhutdinov, R., and Tenenbaum, J.~B.
\newblock Human-level concept learning through probabilistic program induction.
\newblock \emph{Science}, 350\penalty0 (6266):\penalty0 1332--1338, 2015.

\bibitem[Li et~al.(2017)Li, Vasile, and Belta]{li2017reinforcement}
Li, X., Vasile, C.-I., and Belta, C.
\newblock Reinforcement learning with temporal logic rewards.
\newblock In \emph{2017 IEEE/RSJ International Conference on Intelligent Robots
  and Systems (IROS)}, pp.\  3834--3839. IEEE, 2017.

\bibitem[Littman et~al.(2017)Littman, Topcu, Fu, Isbell, Wen, and
  MacGlashan]{littman2017environment}
Littman, M.~L., Topcu, U., Fu, J., Isbell, C., Wen, M., and MacGlashan, J.
\newblock Environment-independent task specifications via gltl.
\newblock \emph{arXiv preprint arXiv:1704.04341}, 2017.

\bibitem[Liu et~al.(2017)Liu, Machado, Tesauro, and
  Campbell]{liu2017eigenoption}
Liu, M., Machado, M.~C., Tesauro, G., and Campbell, M.
\newblock The eigenoption-critic framework.
\newblock \emph{arXiv preprint arXiv:1712.04065}, 2017.

\bibitem[Machado et~al.(2017)Machado, Rosenbaum, Guo, Liu, Tesauro, and
  Campbell]{machado2017eigenoption}
Machado, M.~C., Rosenbaum, C., Guo, X., Liu, M., Tesauro, G., and Campbell, M.
\newblock Eigenoption discovery through the deep successor representation.
\newblock \emph{arXiv preprint arXiv:1710.11089}, 2017.

\bibitem[Pnueli(1977)]{pnueli1977temporal}
Pnueli, A.
\newblock The temporal logic of programs.
\newblock In \emph{Foundations of Computer Science, 1977., 18th Annual
  Symposium on}, pp.\  46--57. IEEE, 1977.

\bibitem[Precup(2000)]{precup2000temporal}
Precup, D.
\newblock \emph{Temporal abstraction in reinforcement learning}.
\newblock University of Massachusetts Amherst, 2000.

\bibitem[Puterman(2014)]{puterman2014markov}
Puterman, M.~L.
\newblock \emph{Markov decision processes: discrete stochastic dynamic
  programming}.
\newblock John Wiley \& Sons, 2014.

\bibitem[Saxe et~al.(2017)Saxe, Earle, and Rosman]{saxe2017hierarchy}
Saxe, A.~M., Earle, A.~C., and Rosman, B.
\newblock Hierarchy through composition with multitask lmdps.
\newblock In \emph{Proceedings of the 34th International Conference on Machine
  Learning-Volume 70}, pp.\  3017--3026. JMLR. org, 2017.

\bibitem[Shah et~al.(2018)Shah, Kamath, Shah, and Li]{shah2018bayesian}
Shah, A., Kamath, P., Shah, J.~A., and Li, S.
\newblock Bayesian inference of temporal task specifications from
  demonstrations.
\newblock In \emph{Advances in Neural Information Processing Systems}, pp.\
  3808--3817, 2018.

\bibitem[Singh et~al.(1995)Singh, Jaakkola, and Jordan]{singh1995reinforcement}
Singh, S.~P., Jaakkola, T., and Jordan, M.~I.
\newblock Reinforcement learning with soft state aggregation.
\newblock In \emph{Advances in neural information processing systems}, pp.\
  361--368, 1995.

\bibitem[Stachenfeld et~al.(2017)Stachenfeld, Botvinick, and
  Gershman]{stachenfeld2017hippocampus}
Stachenfeld, K.~L., Botvinick, M.~M., and Gershman, S.~J.
\newblock The hippocampus as a predictive map.
\newblock \emph{Nature neuroscience}, 20\penalty0 (11):\penalty0 1643, 2017.

\bibitem[Sutton et~al.(1999)Sutton, Precup, and Singh]{sutton1999between}
Sutton, R.~S., Precup, D., and Singh, S.
\newblock Between mdps and semi-mdps: A framework for temporal abstraction in
  reinforcement learning.
\newblock \emph{Artificial intelligence}, 112\penalty0 (1-2):\penalty0
  181--211, 1999.

\bibitem[Todorov(2009)]{todorov2009efficient}
Todorov, E.
\newblock Efficient computation of optimal actions.
\newblock \emph{Proceedings of the national academy of sciences}, 106\penalty0
  (28):\penalty0 11478--11483, 2009.

\bibitem[Wilson et~al.(2007)Wilson, Fern, Ray, and Tadepalli]{wilson2007multi}
Wilson, A., Fern, A., Ray, S., and Tadepalli, P.
\newblock Multi-task reinforcement learning: a hierarchical bayesian approach.
\newblock In \emph{Proceedings of the 24th international conference on Machine
  learning}, pp.\  1015--1022. ACM, 2007.

\end{thebibliography}
\bibliographystyle{icml2019}
\end{document}